\documentclass[Afour,sageh,times]{sagej}

\usepackage{moreverb,url}
\usepackage{graphics} 
\usepackage{epsfig} 
\usepackage{mathptmx} 
\usepackage{times} 
\usepackage{amsmath} 
\usepackage{amssymb}  
\usepackage{bbm}
\usepackage{siunitx}
\usepackage{subcaption}
\usepackage{booktabs}
\usepackage{float}
\usepackage{tabularx}
\usepackage{makecell}
\usepackage{multirow}
\usepackage{mathtools}
\usepackage{hyperref} 

\hypersetup{
    colorlinks=true,
    linkcolor=blue,   
    citecolor=blue,   
    urlcolor=blue     
}

\usepackage[nameinlink,noabbrev]{cleveref}

\DeclareMathAlphabet{\mathcal}{OMS}{cmsy}{m}{n}
\DeclareMathOperator{\E}{\mathbb{E}}
\DeclareMathOperator*{\argmin}{arg\,min}
\newcommand{\R}{\mathbb{R}}

\newcommand\BibTeX{{\rmfamily B\kern-.05em \textsc{i\kern-.025em b}\kern-.08em
T\kern-.1667em\lower.7ex\hbox{E}\kern-.125emX}}

\def\volumeyear{2016}

\begin{document}

\runninghead{Llanes et al.}

\title{Merging model-based control with multi-agent reinforcement learning for multi-agent cooperative teaming strategies}

\author{Christian Llanes\affilnum{1,2}, Spencer W. Jensen\affilnum{2}, and Samuel Coogan\affilnum{1}}

\affiliation{\affilnum{1}School of Electrical and Computer Engineering,
       Georgia Institute of Technology, Atlanta, GA 30332, USA.\\
\affilnum{2}Sandia National Laboratories, Albuquerque, NM 87123, USA.}

\corrauth{Christian Llanes, Georgia Tech and Sandia National Labs.}

\email{christian.llanes@gatech.edu, christianllanes96@gmail.com}

\begin{abstract}
In this work, we propose a framework that combines multi-agent reinforcement learning (MARL) with model-based control to achieve safe, dynamically feasible actions in cooperative multi-agent tasks. Multi-agent reinforcement learning provides the advantage of learning cooperative policies for multi-agent teams from discrete non-differentiable rewards in a long planning horizon. Model-predictive control is robust and offers safe, dynamically feasible actions in a fast replanning framework for short horizons. We propose an algorithm that extends actor-critic model predictive control for MARL which we refer to as multi-agent actor-critic model predictive control (MA-AC-MPC). We demonstrate the capabilities of this algorithm by applying it to a multi-agent pursuit-evasion scenario. Specifically, we compare the evader team's strategy using the MA-AC-MPC model and a multi-layer perceptron model (MA-AC-MLP). The pursuer team uses augmented proportional navigation as it is accepted as an advanced adversarial control law. We also provide an example with a heterogeneous environment where a drone and omni-wheeled rover cooperate to achieve repeatable and successful landing with $100\%$ success rate in hardware for MA-AC-MPC compared to $60\%$ for MA-AC-MLP. We demonstrate the robustness of the proposed MA-AC-MPC algorithm in hardware for both environments.
\end{abstract}


\maketitle

\section{Introduction}

\begin{figure}[t!]
  \centering
  \includegraphics[width=\linewidth]{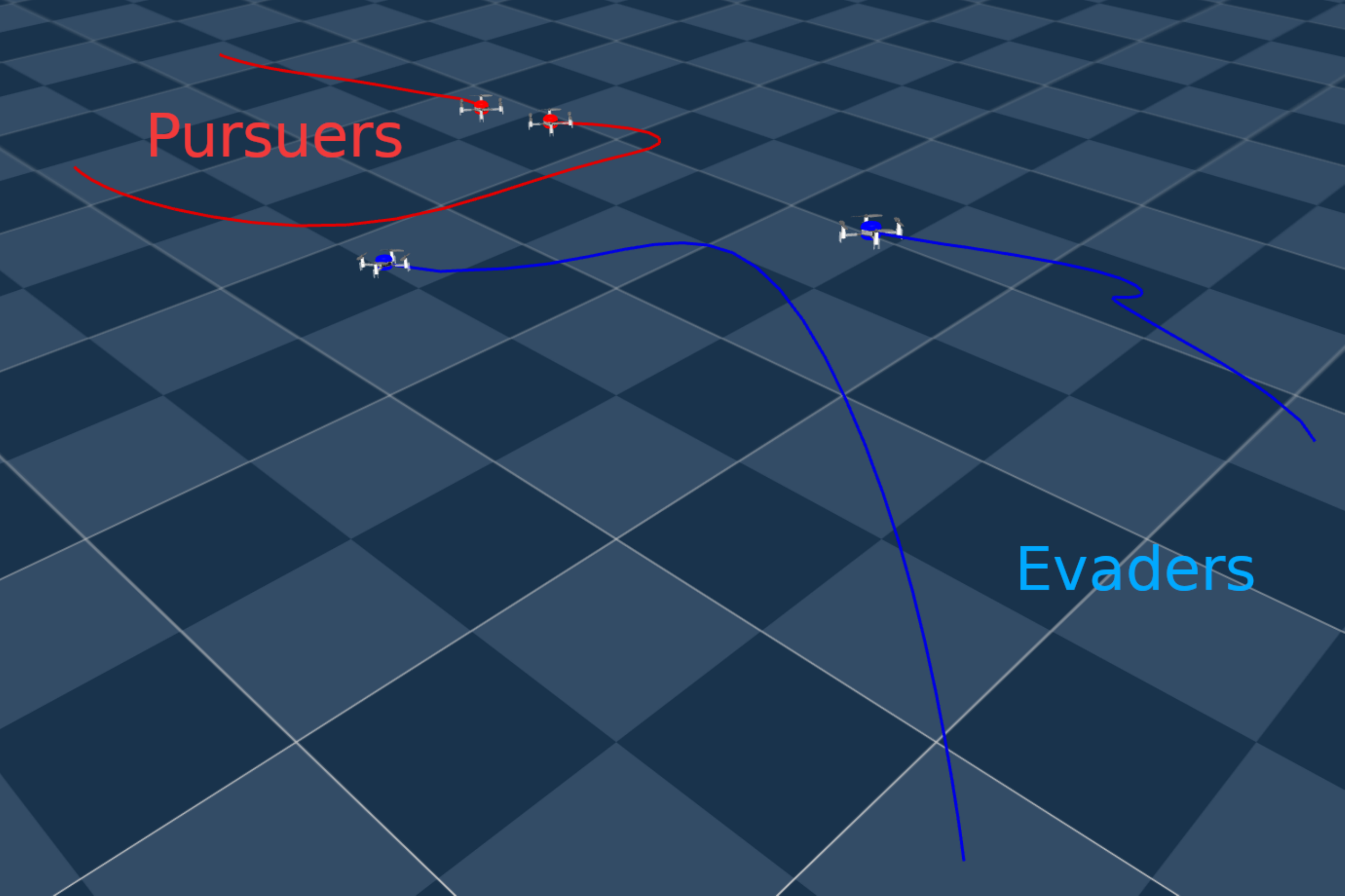}
  \includegraphics[width=\linewidth]{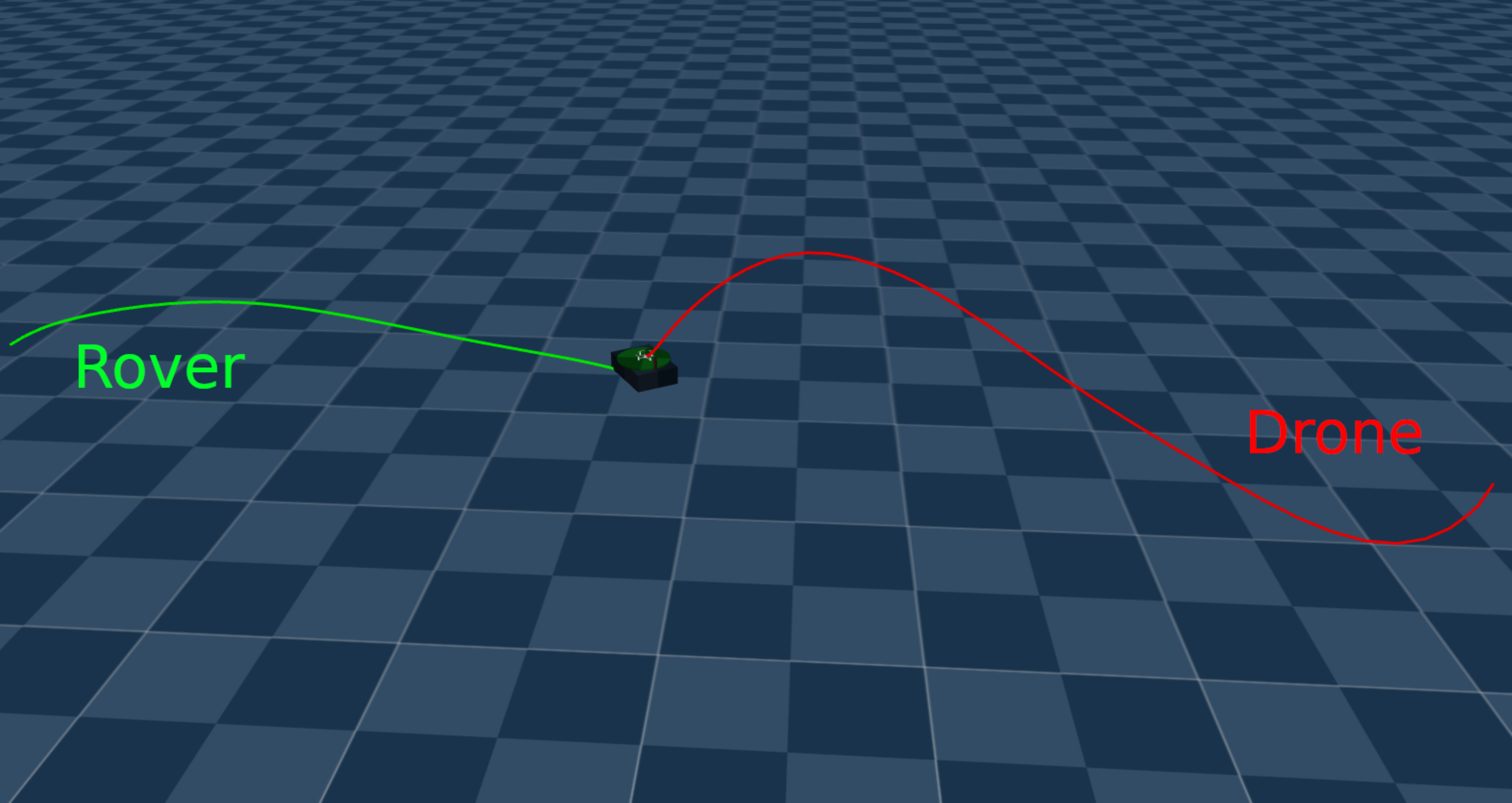}
  \caption{Screenshots of both a homogeneous and heterogeneous multi-agent environment where multi-agent actor-critic model predictive control (MA-AC-MPC) is implemented. The top image is a screenshot at the final step of a two pursuer versus two evader environment trajectory before the pursuers collide into each other. The bottom screenshot is the final step of a drone landing on a rover environment where both agents learn their own AC-MPC policy. Both screenshots are from MuJoCo for rendering.}
  \label{fig:figure 1}
\end{figure}


Multi-agent reinforcement learning (MARL) \citep{4445757} has emerged as a tool for modeling and control of environments with multiple agents acting as decision makers. MARL has been driven by robotics applications where coordination among autonomous agents such as warehouse robots, vehicle fleets, and robotic swarms is essential. Its capabilities have also been demonstrated in developing policies for heterogeneous multi-agent video games with complex tasks and constraints \citep{starcraftmarl}.

However, despite their success, pure learning-based techniques suffer from well-known limitations in maintaining robustness in out-of-distribution scenarios, large sample complexity, and lack of physical real-world constraints. All of which deter the use of pure learning-based methods in safety-critical applications.

Model-based control provides several key advantages by leveraging the known physics and constraints of the robotic application to compute a dynamically feasible and safe action. Model-predictive control (MPC) recomputes the optimization problem over a receding horizon while continuously updating the solution. This enables fast online replanning and enhances robustness. However, this requires that the objective function is differentiable and that the solution be computed quickly enough for the desired robotics application. This also means that the objective function must be manually crafted to fit the desired task which is not feasible for a large array of tasks especially in a multi-agent cooperation problem.

To bridge the gap, we propose combining model-based control and multi-agent reinforcement learning for computing safe dynamically feasible actions that respect the dynamic constraints of each agent, maintain robustness, and are capable of maximizing complex, non-differentiable, and cooperative team rewards. To accomplish this, we extend Actor-Critic Model Predictive Control (AC-MPC) \citep{romero2025actor} for multi-agent reinforcement learning which we refer to as multi-agent actor-critic model predictive control (MA-AC-MPC). To test the capabilities of the MA-AC-MPC we apply it to a multi-agent pursuit-evasion  (MAPE) scenario \citep{Llanes2026ICRA} and compare it to the results from using a multi-agent actor-critic multilayer perceptron (MA-AC-MLP) model.

We summarize our contributions as follows:

\begin{enumerate}
    \item We extend the AC-MPC algorithm to multi-agent reinforcement learning problems with cooperating agent teams working together to maximize their total rewards.
    \item We evaluate the MA-AC-MPC algorithm across two distinct benchmarks: 
    (i) a 2 vs.~2 multi-agent pursuit-evasion scenario, where evaders use MA-AC-MPC to induce pursuer collisions before reaching a goal; and 
    (ii) a cooperative landing scenario where a drone and a rover must synchronize their maneuvers to achieve a successful landing on the moving rover platform.
    \item We implement MA-AC-MPC using the open-source project, \texttt{leap-c} \citep{leonard_fichtner_2025_17244101}, which is faster than the previously used differentiable MPC module, \texttt{mpc.pytorch}, in AC-MPC \citep{romero2025actor} and works for general nonlinear optimal control problems with general nonlinear inequality constraints by using \texttt{acados} \citep{acados}.
    \item We compare our MA-AC-MPC approach with an MA-AC-MLP model to demonstrate improved robustness by incorporating MPC as an actor layer. This is done by comparing the success rate of tasks for different variations of independent agent masses between MA-AC-MPC and MA-AC-MLP. We additionally provide an example where MA-AC-MPC resulted in higher success rate in experiments compared to MA-AC-MLP.
\end{enumerate}

\section{Related Works}

The literature on MPC and RL can be generally categorized into three approaches \citep{reiter2025synthesismodelpredictivecontrol}: using MPC as an expert policy, using MPC in the critic, and using MPC directly in the policy. Methods that use MPC as an expert policy typically use some form of imitation learning \citep{plannetx} to match the MPC policy exactly or use the possibly suboptimal MPC to guide the RL exploration \citep{schulz2024learning}. Some methods use MPC as an expert critic \citep{ghezzi2023imitatiompcqloss, anand2023ddpglearningmpc} which allows for defining the constraint satisfaction in the objective. Methods that use MPC directly in the policy either learn a parametrized dynamical model \citep{lambert2019lowlevelcontrolmodelbasedrl}, the stage cost to encode complex task-specific behavior  \citep{song2022policysearchmpc, difftune, romero2025actor}, or the terminal value function \citep{MORENOMORA20233874, reiter2025ac4mpc} which relates more to stability guarantees. Some methods also use MPC directly as an optimization layer in the actor policy during RL training which results in a closed-loop or end-to-end learning approach \citep{romero2025actor}. This approach can require computing the gradients of the MPC during back-propagation. Several differentiable MPC methods have been developed \citep{amos2018diffmpc, Oshin-RSS-24} with a tradeoff between gradient accuracy and computation time. 

Literature that embeds RL with MPC is typically focused on single-agent RL problems while literature on the multi-agent RL with MPC problem is limited. \cite{MalAir:24-012} proposes a multi-agent RL algorithm that uses distributed MPC for each agent that contributes to the global value function in a centralized learning framework. However, learning is not done in an end-to-end framework which limits its capabilities for learning complex behaviors efficiently such as in cooperation. We contribute to this line of research by extending the single-agent AC-MPC to multi-agent environments.

\section{Preliminaries}
\subsection{Decentralized Markov Decision Processes}
A multi-agent system defines a collection of agents interacting jointly within a shared environment, collaboratively or competitively, to meet some objectives. Often, this requires some form of communication and coordination between agents to complete a shared task. Communication can be in the form of a complete observation of the states of other agents and the global state of the environment or a partial observation with only public states of agents being shared while private states being hidden. For decentralized agents with partial observability the Markov decision process (MDP) is known as a decentralized partially observable Markov decision process (dec-POMDP) \citep{decpomdp} defined by the tuple
\begin{equation}
    M := (\mathcal{K}, \mathcal{S}, \mathcal{A}, P, \mathcal{O}, O, R, \gamma)
    \label{eq:decpomdp}
\end{equation}
where $\mathcal{K} = \{1,..,K\}$ is the set of $K>1$ agents, $\mathcal{S}$ is the global state space of the environment, and $\mathcal{A} = \{\mathcal{A}^{(i)}\}_{i=1}^K$ is the set of action spaces from which an agent can select an action $a^{(i)} \in \mathcal{A}^{(i)}$. Given the global state $s_{t} \in \mathcal{S}$ and the set of agent actions $\mathbf{a} = \{ a^{(1)}, ..., a^{(K)} \}$, then the probability of transitioning to the next state $s_{t+1}$ is defined by the state transition probability function $P(s_{t+1}\mid s_{t}, \mathbf{a})$. $\mathcal{O} = \{\mathcal{O}^{(i)}\}_{i=1}^K$ is the set of individual agent observation spaces $\mathcal{O}^{(i)}$. Each agent receives an observation $o^{(i)}_{t} \in \mathcal{O}^{(i)}$ from  $\mathbf{o}_t = \{o^{(1)}_t,...,o^{(K)}_{t}\}$ given the observation probability function $O(\mathbf{o}_{t+1}\mid \mathbf{a}_{t}, s_{t+1})$. The immediate reward function for each agent is defined by $R^{(i)}(s_{t},\mathbf{a}_{t},s_{t+1})$ with a discount factor of $\gamma \in [0,1)$. In a shared reward \citep{jiang2024fullydecentralizedcooperativemultiagent} or fully cooperative setting \citep{Gronauer2022} the reward is equally shared between all agents $R = R^{(i)} = R^{(K)}$.  

The actions of the individual agent are sampled from a decentralized policy $\pi^{(i)}(a^{(i)} \mid o^{(i)})$. The objective is to find a joint policy $\Pi = \{\pi^{(1)},...,\pi^{(K)}\}$ that maximizes the discounted expected cumulative reward
\begin{equation}
    \underset{\Pi}{\text{maximize }} \E_{s_{t+1} \thicksim P, \mathbf{a}_t \thicksim \Pi} \bigg[ \sum_{t=0}^{\infty}\gamma^t \sum_{i=1}^{K}R^{(i)}(s_t,\mathbf{a}_t, s_{t+1}) \bigg].
    \label{eq: decpomdp_objective}
\end{equation}
For a team of collaborating homogeneous agents, all agents are of the same type and can therefore use a shared policy network that is executed in a distributed manner with each agent acting on its own local observation. Parameter sharing is an effective way \citep{gupta2017} to speed up the learning process when agents have a common goal and need to learn similar behaviors.
Synthesizing a policy to solve \eqref{eq: decpomdp_objective} can be a challenging task. The first reason is that if we consider agents to learn independently and observe other agents as part of the environment, then the environment becomes non-stationary in the perspective of each agent and will be harder to learn a policy that converges to an optimal solution. This is known as fully decentralized learning and execution. An approach to improve the learning of cooperative tasks between agents is incorporating centralized training \citep{Gronauer2022}. This is typically done by including additional global state information in the learning phase. If the additional global state information is discarded and only local agent information is available at execution time, then this is known as centralized training with decentralized execution (CTDE) \citep{jiang2024fullydecentralizedcooperativemultiagent}. If a joint policy is available for all units collectively and they all fully share information with each other, then this is known as centralized training and execution (CTE) \citep{amato2024introductioncentralizedtrainingdecentralized}. However, inter-agent communication can be expensive, particularly when bandwidth or latency limitations are present. One way to address this is to learn a communication protocol between agents \citep{commnet2016, dial2016}. Additional solutions \citep{SANTOS2025104404} have been proposed for dropout prone communication networks where the communication level is not known at the execution time.

For a team of collaborating homogeneous agents, all agents are of the same type and can therefore use a shared policy network that is executed in a distributed manner with each agent acting on its own local observation. Parameter sharing is an effective way \citep{gupta2017} to speed up the learning process when agents have a common goal and need to learn similar behaviors.

\subsection{Nonlinear Model Predictive Control}
For a given discrete-time dynamical system $x_{k+1} = f_k(x_k, u_k;\theta_{\text{MPC}})$ with state $x_k \in \mathcal{X}$, control $u_k \in \mathcal{U}$, and parameters $\theta_{\text{MPC}}$ we define a nonlinear MPC formulation as an optimal control problem (OCP)
\begin{equation}
\begin{split}
\pi_{\text{MPC}}(x;\theta_{\text{MPC}}) &= \argmin_{u} 
\sum_{k=0}^{N-1}  L_k(x_k, u_k; \theta_{\text{MPC}}) + M(x_N;\theta_{\text{MPC}}) \\
\text{s.t.} \quad x_0 &= x \\
x_{k+1} &= f_k(x_k, u_k; \theta_{\text{MPC}}), \quad k = 0, \dots, N-1\\
0 &\geq h_k(x_k, u_k;\theta_{\text{MPC}}), \quad k = 0, \dots, N-1 \\
0 &\geq h_N(x_k;\theta_{\text{MPC}})
\label{eq: MPC}
\end{split}
\end{equation}
where the state and control constraints are contained in the path constraint function $h_k(x_k, u_k;\theta_{\text{MPC}})$ and the terminal constraint function $h_N(x_k;\theta_{\text{MPC}})$. The cost function for this optimal control problem consists of the stage cost $L_k(x_k, u_k; \theta_{\text{MPC}})$ and the terminal cost $M(x_N;\theta_{\text{MPC}})$. 

In a trajectory tracking MPC formulation, the cost function may be written in a nonlinear least-squares form, such as
\begin{equation}
\begin{split}
L_k(x_k, u_k; \theta_{\text{MPC}}) &= \frac{1}{2}
\lVert  y_k(x_k, u_k; \theta_{\text{MPC}}) - y_{k,\text{ref}}(\theta_{\text{MPC}})\rVert^2_{W} \\
M(x_N;\theta_{\text{MPC}}) &= \frac{1}{2}\lVert   y_N(x_N; \theta_{\text{MPC}}) - y_{N,\text{ref}}(\theta_{\text{MPC}})\rVert^2_{W_N} 
\end{split}
\end{equation}
for stage reference $y_{k,\text{ref}} \in \R^{n_y}$, terminal reference $y_{N,\text{ref}} \in \R^{n_{y,N}}$, and weighting matrices $W \in \R^{n_y \times n_y}$, $W_N \in \R^{n_{y,N} \times n_{y,N}}$ from the weighted L2-norm, i.e.
$\lVert x \rVert^2_{W} = x^T W x$. This formulation can also be written as linear least-squares through a linear transformation for $y_k(x_k, u_k; \theta_{\text{MPC}})$ and $y_N(x_N; \theta_{\text{MPC}})$. The cost function may also be written as a quadratic cost function such as
\begin{equation}
L_k(x_k, u_k; \theta_{\text{MPC}}) = \frac{1}{2}
\begin{bmatrix} x_k \\ u_k \end{bmatrix}^\top 
Q_k(\theta_{\text{MPC}})
\begin{bmatrix} x_k \\ u_k \end{bmatrix} + 
p_k(\theta_{\text{MPC}})^\top
\begin{bmatrix} x_k \\ u_k \end{bmatrix},
\label{eq:quadratic_cost_function}
\end{equation}
where $Q_k(\theta_{\text{MPC}})$ and $p_k(\theta_{\text{MPC}})$ denote the quadratic and linear cost coefficients at prediction stage $k$, respectively. The vector $\theta_{\text{MPC}}$ contains the learned parameters that determine these cost coefficients. As shown above, parameters may appear in the cost, dynamics, or constraints. There can also be global parameters such as mass, moment of inertia, spring constants, or fixed cost function coefficients. By defining both the system dynamics and the cost function as neural networks, we treat their underlying weights and biases as tuneable parameters. Adjusting the parameters may lead to improved real-world tracking performance through manual tuning or gradient descent techniques. To perform gradient descent, we require the computation of the sensitivity of the MPC optimal solution with respect to the parameter set. Consequently, this necessitates a differentiable MPC framework that allows gradients to propagate through the optimization solver. Additionally, the differentiable MPC framework should work with existing learning-based libraries such as PyTorch. \cite{amos2018diffmpc} proposed a differentiable MPC framework called \texttt{mpc.pytorch} that computes gradients of a loss function with respect to the MPC parameters by implicitly differentiating KKT conditions at a fixed point of a box-constrained iterative LQR solver. However, their implementation is restricted to quadratic cost functions such as in \eqref{eq:quadratic_cost_function} and box constraints on the control input. Other differentiable optimal control frameworks have been proposed \citep{dinev2022differentiableoptimalcontroldifferential, adabag2025differentiablemodelpredictivecontrol, jin2021pontryagindifferentiableprogrammingendtoend, 10384034}, however, their solution sensitivities are often degraded or restricted to unconstrained optimal control problems. Additionally, the above approaches cannot handle general nonlinear parametric inequality constraints with states and controls. The open-source software framework \texttt{acados} \citep{acados} is  designed for optimal control problems with nonlinear and parametric costs, dynamics, and inequality constraints with the speed required for real-time control applications. Sensitivity propagation has also been implemented in the \texttt{acados} framework \citep{Frey2023, frey2025differentiablenonlinearmodelpredictive} for computing derivatives of the optimal solution with respect to parameters and initial state. Recently, the open-source project \texttt{leap-c} has provided differentiable interfaces for the \texttt{acados} sensitivity implementation within machine learning frameworks such as \texttt{PyTorch} and \texttt{JAX}. \citep{leonard_fichtner_2025_17244101}.
\begin{figure*}[t!]
  \centering
  \includegraphics[width=\linewidth]{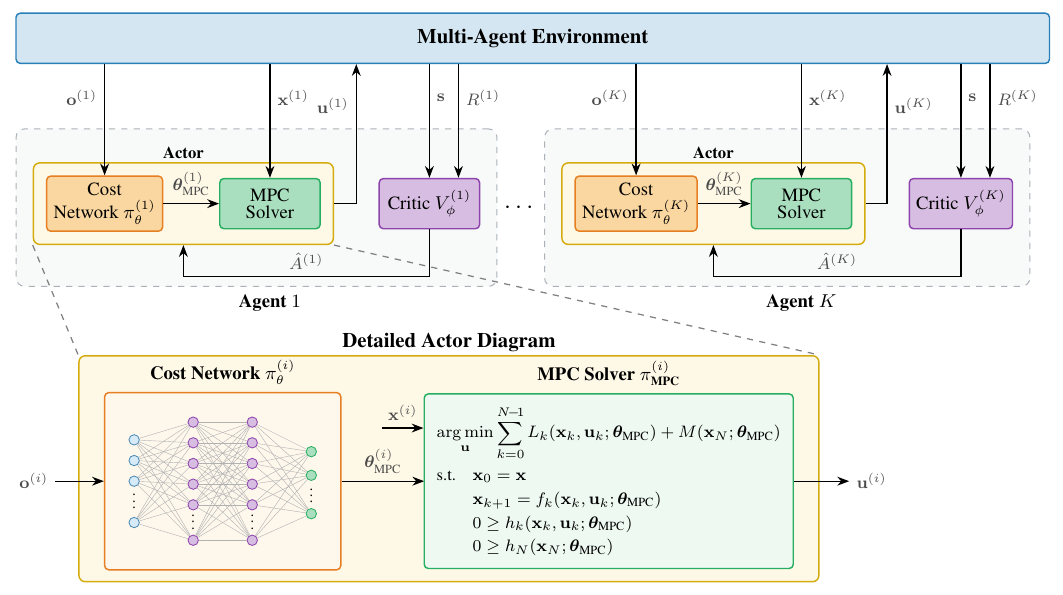}
  \caption{Architecture diagram for multi-agent actor-critic model predictive control (MA-AC-MPC).}
  \label{fig: architecture diagram}
\end{figure*}

\section{Methodology}
In this section, we discuss our implementation of actor-critic model predictive control for multi-agent reinforcement learning problems.
\subsection{Actor-Critic Multi-Agent Reinforcement Learning}
In order to find a joint policy $\Pi = \{\pi^{(1)},...,\pi^{(K)}\}$ that maximizes the discounted expected cumulative returns from \eqref{eq: decpomdp_objective} we turn towards the actor-critic structure in reinforcement learning. To approximate the true value function,
\begin{equation}
\begin{split}
    V_{\pi}(s) =  \E \bigg[ &\sum_{t=0}^{T-1}\gamma^t R(s_t,\mathbf{a}_t, s_{t+1}) \bigm\vert s_0 = s\bigg] \\
    &s_{t+1} \thicksim P, \mathbf{a}_t \thicksim \Pi
\end{split}
\end{equation}
we use a feed-forward neural network that computes a centralized value function for the multi-agent scenario during training given the full state of the environment which may include information not provided at execution time to the policy. To train the actor-critic structure, we use multi-agent proximal policy optimization (MAPPO) \citep{yu2022surprising}. In MAPPO, the general actor loss for potentially heterogeneous agents is defined by the sum of the clipped surrogate objective and the entropy loss
$L^{(i)}_{\text{actor}}(\theta^{(i)}) = L^{(i)}_{\text{clip}}(\theta^{(i)}) + \lambda_{\text{entropy}}L^{(i)}_{\text{entropy}}(\theta^{(i)})$ where the hyperparameter $\lambda_{\text{entropy}}$ is the coefficient of entropy loss. The clipped surrogate objective is defined as
\begin{equation}
\begin{split}
    L^{(i)}_{\text{surrogate}}(\theta^{(i)}) &= r^{(i)}_{\theta, b} \hat{A}^{(i)}_b \\
    L^{(i)}_{\text{clipped surrogate}}(\theta^{(i)}) &= \text{clip}(r^{(i)}_{\theta, b}, 1-\epsilon_{a}, 1+\epsilon_{a})\hat{A}^{(i)}_b \\
    L^{(i)}_{\text{clip}}(\theta^{(i)}) =  -\frac{1}{B} \sum_{b=1}^{B} \Big[ &\min\Big(L^{(i)}_{\text{surrogate}},  L^{(i)}_{\text{clipped surrogate}}\Big) \Big]    
\end{split}
\end{equation}
for batch of size $B$, PPO clipping term $\epsilon_{a}$, and $r^{(i)}_{b}(\theta^{(i)}) = \frac{\pi^{(i)}_\theta(a^{(i)}_b | o^{(i)}_b)}{\pi^{(i)}_{\theta_\text{old}}(a^{(i)}_b | o^{(i)}_b)}$. The clipped surrogate objective updates the actor in a way that maximizes the probability that actions lead to better rewards while clipping the update to stabilize the learning process. 
The advantage estimate $\hat{A}^{(i)}_b$ is calculated using the generalized advantage estimate (GAE) \citep{schulman2018highdimensional} where $\gamma_{\text{GAE}}$ is the reward discount factor and $\lambda_{\text{GAE}}$ is the smoothness factor. The second term in the actor loss function is the entropy loss defined as
\begin{equation}
    L^{(i)}_{\text{entropy}}(\theta^{(i)}) = -\frac{1}{B} \sum_{b=1}^{B} \pi^{(i)}_{\theta_\text{entropy}}(o^{(i)}_b)
\end{equation}
which prevents getting stuck in a local optimum and reinforces exploration. The critic loss function is defined as
\begin{equation}
\begin{split}
    V^{(i)}_{\phi,\text{clipped}}(s_b) &= V^{(i)}_{\phi_{\text{old}}}(s_{b}) + \text{clip}(V^{(i)}_{\phi_{\text{new}}}(s_{b}) - V^{(i)}_{\phi_{\text{old}}}(s_{b}), -\epsilon_{c}, \epsilon_{c})\\
    L^{(i)}_{\text{critic}}(\phi^{(i)}) &= \frac{1}{B} \sum_{b=1}^{B} (V^{(i)}_{\phi,\text{clipped}} - \hat{R}^{(i)}_b)^2
\end{split}
\end{equation}
for critic clipping term $\epsilon_{c}$ and discounted reward-to-go $\hat{R}^{(i)}_b = \hat{A}^{(i)}_b + V^{(i)}_{\phi_{\text{old}}}(s_{b})$. The total loss function is defined as $L(\Theta,\Phi) = \sum_{i=1}^K L^{(i)}_{\text{actor}}(\theta^{(i)}) + \lambda_{\text{critic}}L^{(i)}_{\text{critic}}(\phi^{(i)})$ for $\Theta = \{\theta^{(1)},...,\theta^{(K)}\}$ and $\Phi = \{\phi^{(1)},...,\phi^{(K)}\}$ where the hyperparameter $\lambda_{\text{critic}}$ is the critic loss coefficient. For homogeneous agent environments with shared actor networks the loss function can be reduced to $L(\theta,\phi) = \frac{1}{K}\sum_{i=1}^K L^{(i)}_{\text{actor}}(\theta) + \lambda_{\text{critic}}L^{(i)}_{\text{critic}}(\phi)$. The  parameters $\Theta$ and $\Phi$ are updated via gradient descent on the joint loss function $L(\Theta, \Phi)$. To ensure training stability, the gradient norm is clipped to a maximum threshold of $\lambda_{\text{grad}}$, so that the update follows
\begin{equation}
\hat{g} = \begin{cases} g & \text{if } \|g\|_2 \leq \lambda_{\text{grad}} \\ \frac{\lambda_{\text{grad}}}{\|g\|_2} g & \text{if } \|g\|_2 > \lambda_{\text{grad}} \end{cases}  
\end{equation}
where $g = \nabla_{\Theta,\Phi} L(\Theta, \Phi)$.

In certain multi-agent environments, agents may be terminated or deactivated mid-episode, often as a consequence of events such as collisions. It has been shown by \cite{yu2022surprising} that updating the actor and critic while an agent is inactive without any changes to the agent observation or global state results in poor training results. This leads us to the concept of \textit{death masking} in which an inactive or dead agent has its observation or states masked by a zero vector. \cite{yu2022surprising} proposed an explanation in which a distribution shift occurs in the input of the value network when an agent becomes deactivated, but containing it to a single  vector of zeros allows the value network to better reason about the future rewards of the environment with deactivated agents in the zero vector input. Additionally, they achieved better results by also including an agent identifier with death masking. Therefore, in our implementation we use GAE bootstrap cutoff for inactive agents, masking inactive agent states and observations, one-hot encoded agent ID, and a binary flag indicating whether an agent is active.

\subsection{Augmenting MARL with MPC}

The MPC problem defined in \eqref{eq: MPC} is defined as a single-agent MPC problem. Transforming this problem as a multi-agent MPC problem can we done in several ways. One method is to have a centralized MPC solver that controls agents externally. However, this approach scales poorly with the number of agents as the number of decision variables is increased and is not effective for autonomous environments where a centralized computer is required. Additionally, this approach requires reliable communications as control actions are published through the network. Even then, latency remains a problem, limiting the effective capabilities of each agent. A secondary approach is a distributed MPC formulation where a centralized problem can be distributed into several smaller subsystems \citep{negenborn2009multiagentmodelpredictivecontrol} that communicate with each other. Agents that do not interact with each other are known as decentralized agents.

The challenge in designing a distributed MPC framework is constructing a proper objective for each agent that formalizes the general cooperative task. Additionally, the objective and constraints must remain tractable enough to compute the optimal control in real-time. In the context of \eqref{eq: MPC} the cooperative task must be encoded through the MPC parameters $\theta_{\text{MPC}}$. More specifically, the stage and terminal cost are designed for some cooperative objective.  In our approach, we use a linear least-squares formulation for the cost function where the tuneable parameters are the stage reference $y_{k,\text{ref}}$, the terminal reference $y_{N,\text{ref}}$, and the weighting matrices $W$, $W_N$. More formally, we use an AC-MPC structure with a hierarchical actor such that the first layer is a neural network that outputs the MPC parameters $\pi^{(i)}_{\theta}(y^{(i)}_{k,\text{ref}},y^{(i)}_{N,\text{ref}},W^{(i)},W^{(i)}_{N} | o^{(i)})$ which behaves as more of a planner and the second layer is the differentiable MPC module using \texttt{leap-c} to generate feasible control actions. The deployment is distributed, however, the neural network cost function takes observations that contain information about the agent's own state and partial state information from other agents. During the centralized training phase, a shared value function evaluates the global environment state to optimize individual actor policies toward maximizing collective rewards. An architecture diagram of MA-AC-MPC is provided in \Cref{fig: architecture diagram}.

\section{Results}
In this section, we evaluate the proposed MA-AC-MPC framework on two different multi-agent environments and compare the results to a standard multilayer perceptron actor. The first environment is a multi-agent pursuit evasion environment with two pursuers and two evaders. The evaders use MA-AC-MPC and the pursuers use a fixed pursuit strategy. The second environment is a drone and rover environment where the drone is tasked with landing on a rover and both agents have their own policy and MPC model. This allows us to compare MA-AC-MPC for both a homogeneous environment with a shared policy for each agent and a homogeneous environment with different policies for each agent. For training our models, we use SKRL \citep{skrl} with its in-built implementation of the MAPPO algorithm and a slight modification to include MPC initial states in the memory buffer.

\begin{table}[t]
\small\sf\centering
\caption{Reward function components and coefficients.}
\label{tab: reward mape}
\begin{tabular}{@{}llr@{}}
\toprule
Component & Expression & Coefficient \\
\midrule
\multicolumn{3}{@{}l}{\textit{Sparse event rewards}} \\
Capture             & $r_\text{cap} \cdot \sum_{i=1}^{N_e} \mathbbm{1}[\text{capture}_{e_i}]$                                   & $-5.0$ \\
Evader collision    & $r_\text{ee} \cdot \sum_{i=1}^{N_e} \mathbbm{1}[\text{collision}_{e_i}]$                                  & $-5.0$ \\
Pursuer collision   & $r_\text{pp} \cdot \sum_{i=1}^{N_p} \mathbbm{1}[\text{collision}_{p_i}]$                                  & $30.0$ \\
PP closing velocity & $c_\text{rv} \cdot \sum_{i \neq j} \max(-\mathbf{v}_{ij}^\top \hat{\mathbf{r}}_{ij},\; 0)$                & $1.0$ \\
Boundary violation  & $r_\text{bnd} \cdot \sum_{i=1}^{N_e} \mathbbm{1}[\text{geofence}_{e_i}]$                                  & $-5.0$ \\
\addlinespace
\multicolumn{3}{@{}l}{\textit{Shaping rewards}} \\
Pursuer proximity   & $c_\text{prox} \cdot e^{-\lambda_\text{prox} \, \lVert \mathbf{p}_{p_1} - \mathbf{p}_{p_2} \rVert}$      & $0.3,\;\lambda{=}2.5$ \\
\addlinespace
\multicolumn{3}{@{}l}{\textit{Penalties (subtracted)}} \\
Attitude            & $c_\Phi \cdot \sum_{i=1}^{N_e} \lVert \boldsymbol{\Phi}_{e_i} \rVert$                                     & $0.01$ \\

Velocity            & $c_v \cdot \sum_{i=1}^{N_e} \lVert \mathbf{v}_{e_i} \rVert^2$                                            & $0.03$ \\
Hover thrust     & $c_u \cdot \sum_{i=1}^{N_e} (u_{e_i}^{T} - u_\text{hover}^{T})^2$                                        & $0.5$ \\
Thrust smoothness   & $c_{\Delta u_T} \cdot \sum_{i=1}^{N_e} (\Delta u_{e_i}^{T})^2$                                           & $5.0$ \\
Control smoothness  & $c_{\Delta u} \cdot \sum_{i=1}^{N_e} \lVert \Delta \mathbf{u}_{e_i}^{\text{rpy}} \rVert^2$               & $1.0$ \\
\bottomrule
\end{tabular}
\vspace{2pt}
\begin{flushleft}
$\mathbbm{1}[\cdot]$ denotes the indicator function. All terms are summed over active agents and normalized by $N_\text{pairs} = 2$.
\end{flushleft}
\end{table}

\subsection{Multi-Agent Pursuit Evasion}
\subsubsection{Problem Definition}
A multi-agent pursuit-evasion scenario was proposed by \cite{Llanes2026ICRA} where a team of evaders and a team of pursuers operate in a 3-dimensional environment. The pursuers use one of two fixed strategies to chase and capture evaders: pure pursuit or augmented proportional navigation \citep{Zarchan2019}. The evaders are tasked with evading the pursuers and are rewarded with reaching a goal with minimal velocity. This environment is a good application for MA-AC-MPC since evaders are assumed to be homogeneous agents with the same dynamics and overall must achieve a goal that requires cooperation. 

We apply this problem for a 2 pursuer versus 2 evader environment with some slight modifications to the original paper to demonstrate that MA-AC-MPC is more robust and sample efficient in learning for this problem. The first modification is to the objective such that there is no longer a goal and the main task of evaders is to make the pursuers collide with each other. 
However, an evader win condition now requires that all pursuers must be dead and at least one evader is alive. A pursuer win condition occurs when all evaders are dead even if all pursuers are dead as well. An evader agent dies if it is captured, crashes onto the ground, crashes with a teammate, or exits the boundary. A pursuer agent dies if it collides with teammates or captures an evader.

We fix the pursuer strategy to use proportional navigation only without feedback of the evader acceleration. This was done to reduce the amount of information that was passed through the limited bandwidth radio hardware. However, we add an additional line-of-sight acceleration to the pursuers such that there is no need for switching to pure-pursuit below some closure velocity. This leads to a satisfactory pursuit policy that captures evaders with near $100\%$ success. Additionally, each evader and pursuer is spawned within its own box that is nominally spaced with its team member. A distance check is performed to ensure that the spawn location exceeds the team's collision radius by a defined safety margin to ensure that members do not collide immediately on environment reset. 

We extend the evader team rewards for the environment with an exponential proximity for the pursuers, an extension to the reward for pursuer-on-pursuer collision by rewarding higher closing velocity on collision, an angle penalty for evader Euler angles $\Phi_e = [\phi, \theta, \psi]$ to deter large orientation maneuvers, a velocity penalty for flying too quickly, a penalty for thrust difference from hover, and a control smoothness penalty for variations in sequential control actions. Together, these extra rewards ensured a satisfactory training process and simulation-to-hardware. 
\begin{table}[t!]
\caption{Curriculum level parameters. Evader spawn is deterministic at nominal positions for levels~1--9 and randomized ($\Delta_{xyz} = 0.1$\,m) at level~10. Pursuer--pursuer and pursuer--evader collision tolerances are equal ($d_\mathrm{pp} = d_\mathrm{pe}$) at all levels. The advance threshold is a 70\% evader win rate.}
\label{tab: curriculum mape}
\small\sf\centering
\scriptsize
\begin{tabular}{@{}c cc cc cc cc@{}}
\toprule
& \multicolumn{2}{c}{Pursuer Spawn} & \multicolumn{2}{c}{Collision Tol.} & \multicolumn{2}{c}{Domain Rand.} & \multicolumn{2}{c}{Disturbance} \\
\cmidrule(lr){2-3} \cmidrule(lr){4-5} \cmidrule(lr){6-7} \cmidrule(lr){8-9}
Level & $\Delta_{xy}$ & $\Delta_z$ & $d_\mathrm{ee}$ & $d_\mathrm{pp}$ & $\sigma_m$ & $\sigma_J$ & $\sigma_f$ & $\sigma_\tau$ \\
& {[m]} & {[m]} & {[m]} & {[m]} & {[g]} & {[$\!\times\!10^{-6}$]} & {[mN]} & {[$\mu$N\,m]} \\
\midrule
1 & 0.10 & 0.10 & 0.20 & 0.50 & --- & --- & --- & --- \\
2 & 0.25 & 0.25 & 0.30 & 0.40 & --- & --- & --- & --- \\
3 & 0.35 & 0.35 & 0.30 & 0.40 & --- & --- & --- & --- \\
4 & 0.40 & 0.40 & 0.30 & 0.40 & 1.0 & 0.5 & --- & --- \\
5 & 0.40 & 0.40 & 0.30 & 0.40 & 2.0 & 1.0 & --- & --- \\
6 & 0.40 & 0.40 & 0.30 & 0.40 & 2.0 & 1.0 & 5.0 & 50 \\
7 & 0.50 & 0.40 & 0.30 & 0.35 & 2.0 & 1.0 & 5.0 & 50 \\
8 & 0.50 & 0.40 & 0.30 & 0.30 & 2.0 & 1.0 & 5.0 & 50 \\
9 & 0.50 & 0.40 & 0.20 & 0.20 & 2.0 & 1.0 & 5.0 & 50 \\
10$^\dagger$ & 0.50 & 0.40 & 0.20 & 0.20 & 2.0 & 1.0 & 5.0 & 50 \\
\bottomrule
\multicolumn{9}{@{}l@{}}{\footnotesize $^\dagger$Evader spawn randomized: $\Delta_{xyz} = 0.1$\,m.}
\end{tabular}
\end{table}
The rewards are more clearly defined in \Cref{tab: reward mape}.

\subsubsection{Model and Training Configuration}
The dynamics in \citep{Llanes2026ICRA} use a simplified multirotor drone model with an approximation of the rotational dynamics as a first order system for training. The authors demonstrated successful sim-to-real, but the dynamics can be improved to transfer policies with more agile maneuvers. This is why we chose to use \texttt{crazyflow} \citep{schuck2025crazyflow}, a fast parallelizable Crazyflie drone swarm simulator using \texttt{JAX}. For the environment simulation, we use the \verb|first_principles| model, which is a high-fidelity full rigid body simulation of the drone with rotor dynamics, aerodynamic drag, and blade gyroscopic effects. The Mellinger controller in \texttt{crazyflow} is used because it contains the same parameters as the Crazyflie firmware. For the MPC model, we use the \verb|so_rpy| model which is faster with a tradeoff of lower complexity. It requires computing system identification coefficients for the rotational dynamics. For all of our experiments, we use the provided system identification parameters for the Crazyflie thrust upgrade model \verb|cf2x_T350|.
\begin{table}[t!]
\small\sf\centering
  \caption{A comparison of training times, inference times, and neural network sizes for MA-AC-MPC with $N=2$ and MA-AC-MLP methods. All MA-AC-MLP training times are for 4M steps and MA-AC-MPC training times are for 2M steps. Inference times are per agent.}
  \label{tab: method comparison}
  \begin{tabularx}{\linewidth}{Xccc}
    \toprule
    Method & Net Size (A / C) & Train Time & Inference Time \\
    \midrule
    MPC & [256$\times$2 / 256$\times$2] & \qty{162.2}{h} & \qty{1.052 \pm 0.474}{ms}\\
    MLP & [256$\times$2 / 256$\times$2] &  \qty{18.0}{h} & \qty{0.120 \pm 0.063}{ms}\\
    MLP & [512$\times$2 / 512$\times$2] &  \qty{26.8}{h} & \qty{0.140 \pm 0.077}{ms}\\
    MLP & [512$\times$3 / 512$\times$2] &  \qty{30.0}{h} & \qty{0.197 \pm 0.175}{ms}\\
    \bottomrule
  \end{tabularx}
\end{table}
\begin{figure}[t!]
  \centering
  \includegraphics[width=\linewidth]{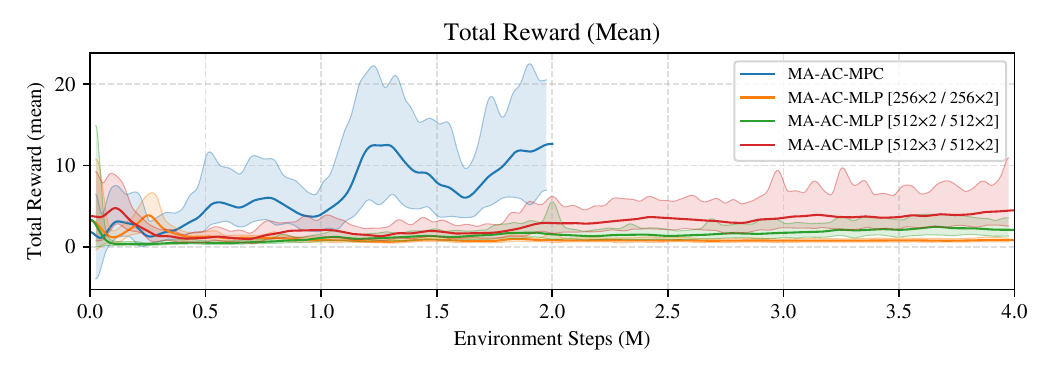}
  \includegraphics[width=\linewidth]{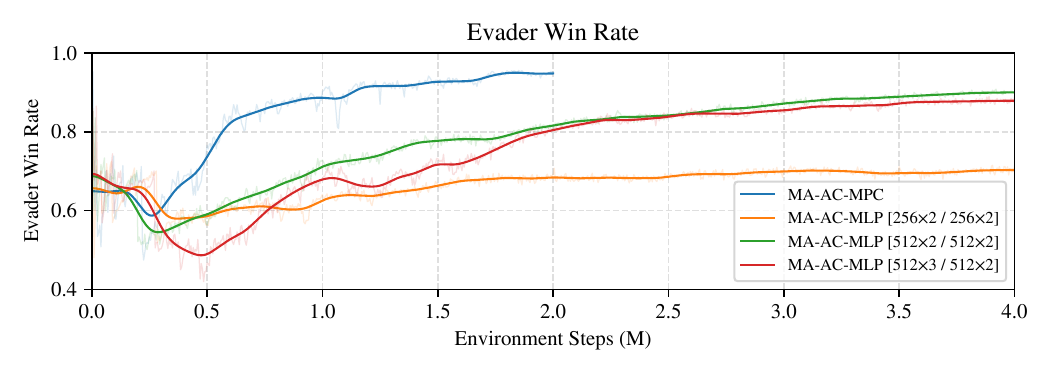}
\includegraphics[width=\linewidth]{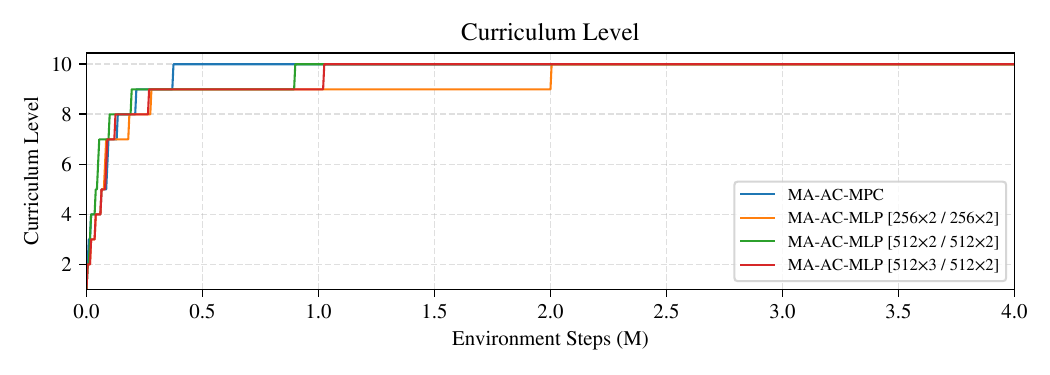}
  \caption{Training curves for mean total reward, evader win rate, and curriculum level for various MA-AC-MLP actor and critic neural network sizes and the proposed MA-AC-MPC. Curriculum level advancement occurs when the evader win rate reaches a threshold of $70\%$.}
  \label{fig: training curves}
\end{figure}
The MA-AC-MPC model uses a single policy shared by both evader agents. The observation to the policy is the agent's position, velocity, rotation matrix flattened, body rates, and a one-hot ID for the agent using the policy. Additionally, the observation includes teammate and pursuer states, such as position and velocity, which are masked based on whether the agent is alive, along with an explicit alive flag. Finally, the target of the pursuer is provided as one-hot vectors. The shared state provided to the critic is similar but includes rotation matrix and body rates for all evaders and pursuers. We trained MA-AC-MPC and different MA-AC-MLP model sizes on a 16-core AMD Ryzen 9 5950x desktop processor and tabulate the training time and inference time in  \Cref{tab: method comparison}. One downside of the \texttt{acados} framework is that the code generation tools are built and optimized for CPU operations and cannot be run on the GPU. The MPC step is $\Delta t=20\unit{ms}$ which results in a control frequency of $50\unit{Hz}$. The horizon for the MPC model $N=2$ was chosen based on the results of \citep{romero2025actor}. Additionally, we tested $N=5$ and did not achieve better training or simulation results. We provide training curves in \Cref{fig: training curves} for MA-AC-MPC and the various MA-AC-MLP model sizes. Although the MA-AC-MPC architecture requires longer training times per iteration compared to the MA-AC-MLP models, it achieves superior win rates in significantly fewer steps. This suggests enhanced representational efficiency; by offloading low-level dynamics to the MPC, the neural network is not forced to encode complex model information. Consequently, the network functions as a high-level task planner rather than a black box, end-to-end controller. 

We initially attempted to train the evader versus pursuer environment with totally random positions, but noticed that the evaders struggled to learn to cooperate to lead the pursuers into a collision. This led us to design a set of curriculum levels with increasingly difficult environment parameters such as larger spawn distance between pursuers, tighter collision tolerances, and more disturbances. These parameters are tabulated in \Cref{tab: curriculum mape} and the progression of levels throughout training is plotted in \Cref{fig: training curves}.

\begin{figure}[t!]
  \centering
  \includegraphics[width=\linewidth]{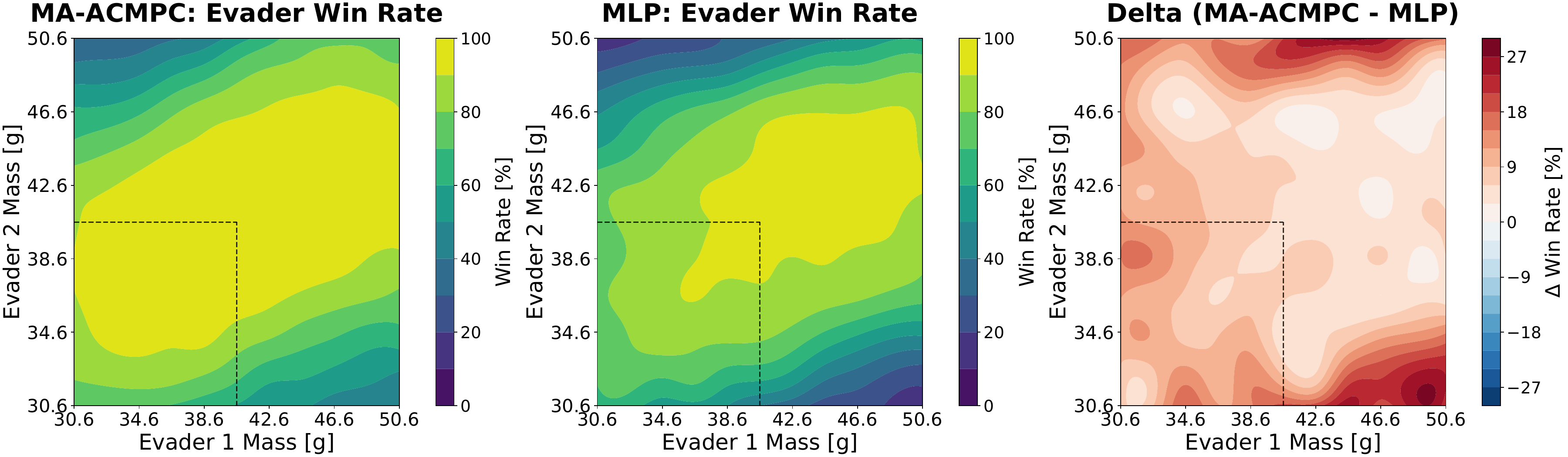}
  \caption{Evader win rate for independent evader mass variations for both MA-AC-MPC [256$\times$2 / 256$\times$2] and MA-AC-MLP [512$\times$2 / 512$\times$2]. Both models are trained with a nominal mass of \qty{40.6}{\gram}. The mass is only changed within the environment and not the MPC dynamics which uses the nominal mass. The third plot is the difference in win rate between both methods where darker red implies MA-AC-MPC win rate is much higher than MA-AC-MLP. This plot is generated for 1000 episodes of each mass variation for each method.}
  \label{fig:mass sweep mape}
\end{figure}
\begin{figure}[t!]
    \centering
    \includegraphics[width=\linewidth]{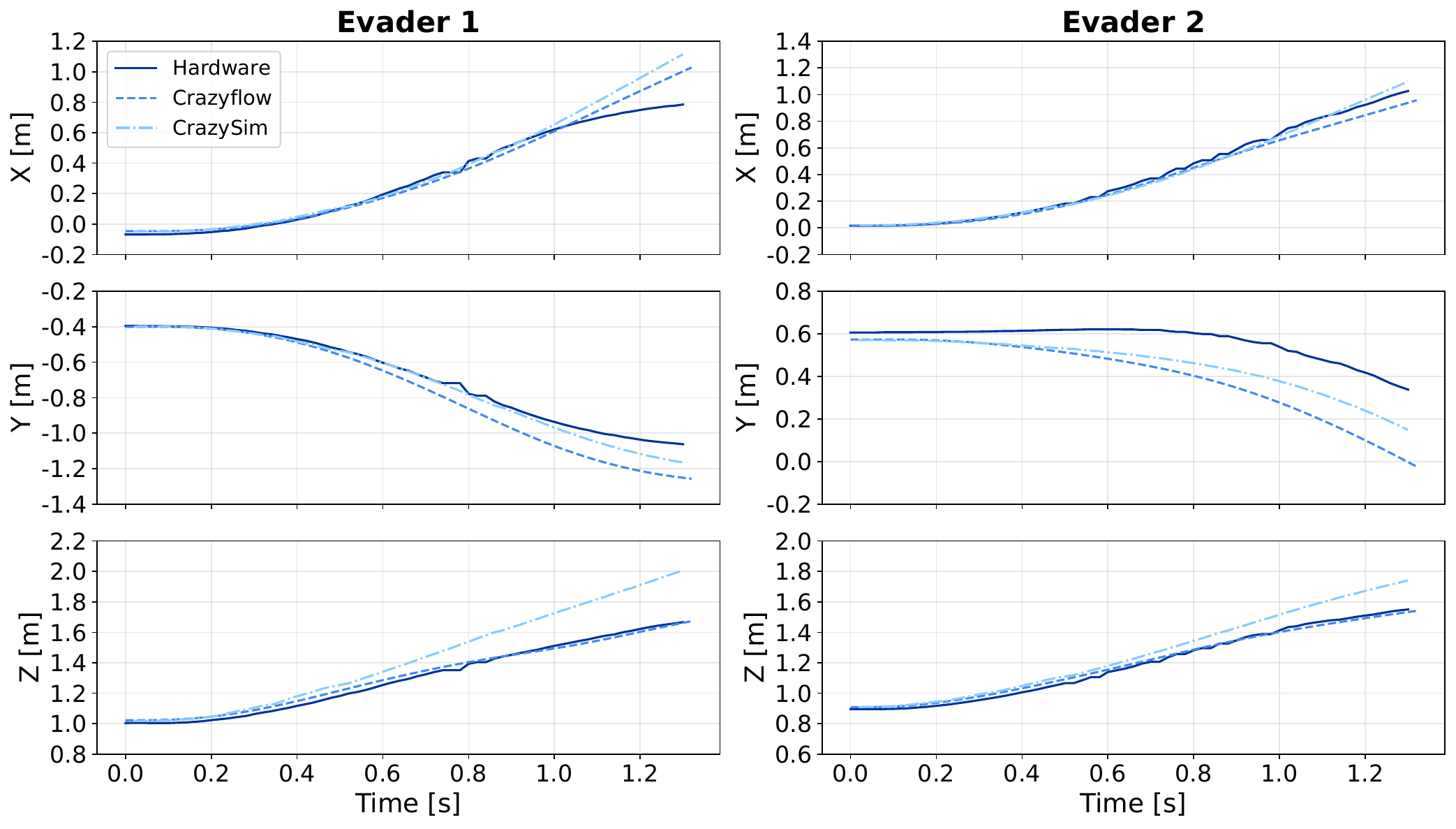}
    \includegraphics[width=\linewidth]{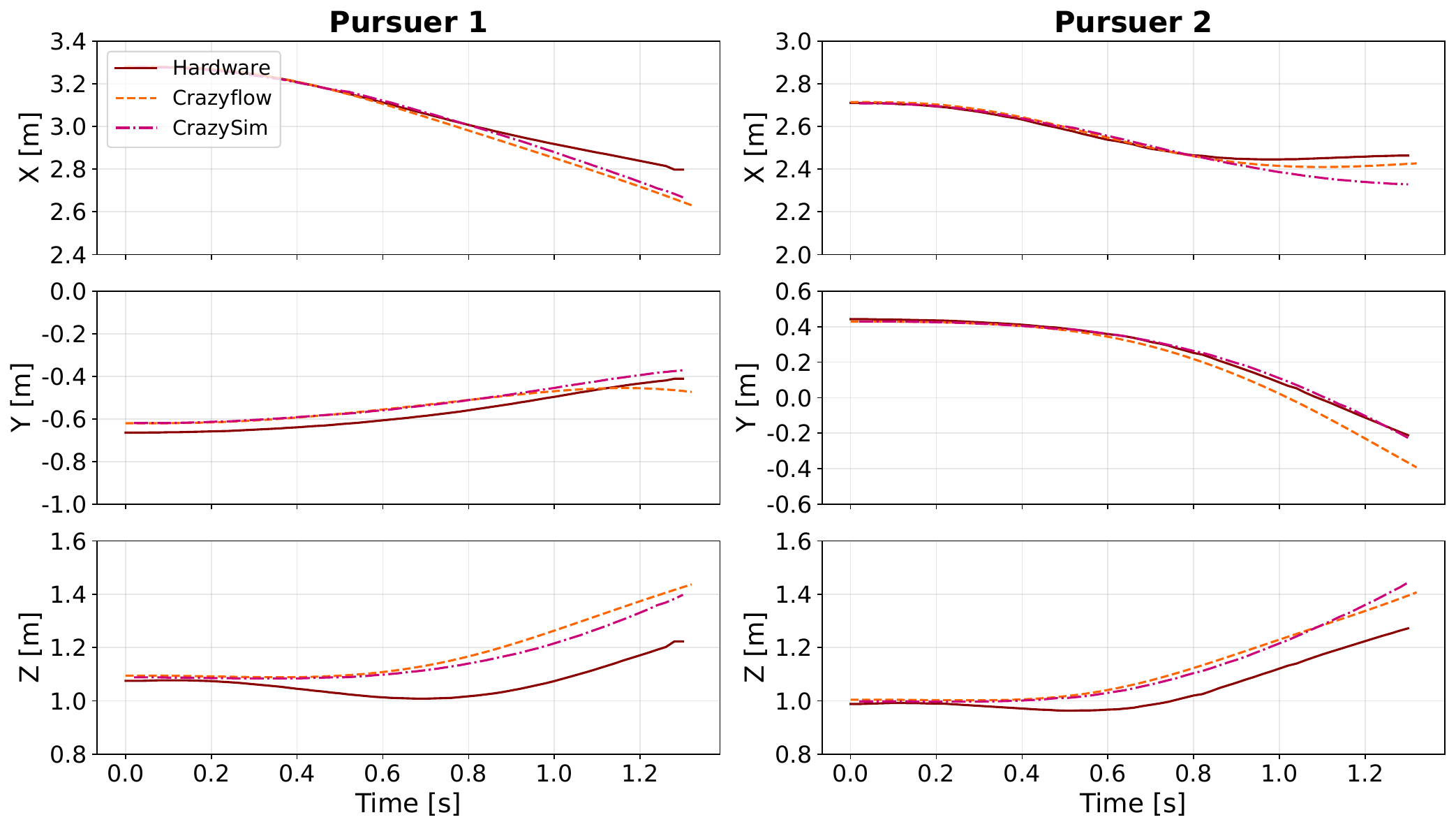}
    \includegraphics[width=\linewidth]{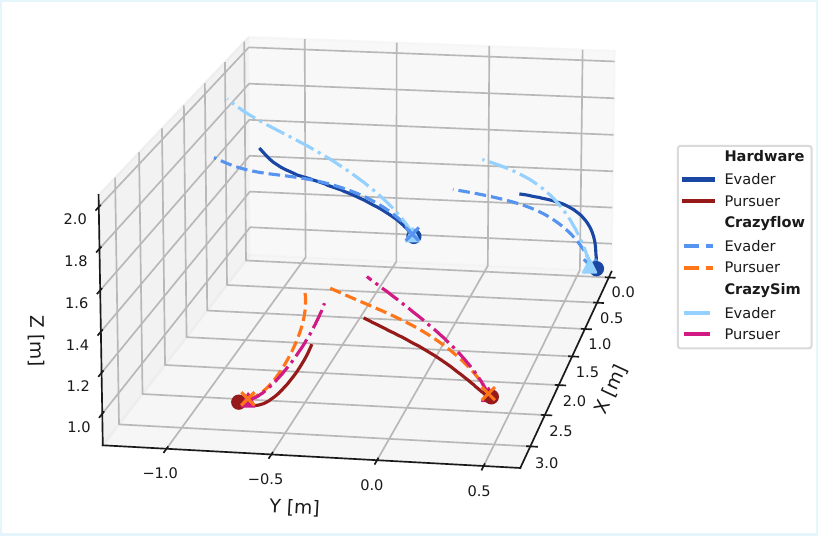}
    \caption{Evader and pursuer position plots for hardware, Crazyflow, and CrazySim. See Extension 1 for Crazyflow simulation video and Extension 2 for hardware video.}
    \label{fig: position plots hardware}
\end{figure}
\subsubsection{Simulation and Hardware Experiments}
We tested MA-AC-MPC trained $(N=2)$ with [Actor size / Critic size] [256$\times$2 / 256 $\times$ 2] against the MA-AC-MLP model size with the highest evader win rate which is [512$\times$2 / 512 $\times$ 2] in a mass sweep robustness test shown in \Cref{fig:mass sweep mape} where the mass is only changed within the environment and not the MPC dynamics. The mass sweep test is performed between \qty{30.6}{\gram} and \qty{50.6}{\gram} in intervals of \qty{2}{\gram} with a nominal trained mass of \qty{40.6}{\gram}. The left plot shows the contour color-coded with evader win rate for the MA-AC-MPC. The middle plot is the evader win rate contour for MA-AC-MLP. The right-most plot is difference in win rate between both models where red implies MA-AC-MPC has higher win rate than MA-AC-MLP. There is no blue shading in the right-most plot implying that MA-AC-MPC outperformed MA-AC-MLP in the mass sweep robustness test.

We deployed the trained MA-AC-MPC model to a ROS 2 framework using Crazyswarm2 \citep{crazyswarm} for hardware testing. However, before hardware, we tested the ROS 2 nodes with the actual Crazyflie firmware in the loop using CrazySim \citep{LlanesICRA2024} with the MuJoCo simulator that models rotor dynamics, aerodynamic effects, and rotor gyroscopic precession in CrazySim. We compared Crazyflow which is the environment used for training, CrazySim, and hardware trajectory data and plotted the results in \Cref{fig: position plots hardware}. The position trajectories for the evaders and pursuers are plotted in the top curves and the 3-dimensional position plot is shown in the bottom. The curves show a similar trend in the trajectory even with hardware sensor noise, disturbances, estimator state uncertainty, and any model mismatch. Another trajectory is shown in \Cref{fig:figure 1} for different initial positions where the evaders cooperate to lead the pursuers into a head-on collision.

\subsection{Drone Landing on Ground Robot}
\begin{table}[t]
\small\sf\centering
\caption{Reward function components and coefficients for the drone and rover landing task.}
\label{tab:reward_landing}
\scriptsize
\setlength{\tabcolsep}{0pt}
\renewcommand{\arraystretch}{1.12}
\begin{tabular}{@{}p{0.28\columnwidth}@{\hspace{2pt}}p{0.5\columnwidth}@{\hspace{2pt}}r@{}}
\toprule
Component & Expression & Coefficient \\
\midrule
\multicolumn{3}{@{}l}{\textit{Team rewards}} \\
XY progress
& $c_\text{prog}(d_\text{xy}^{t-1} - d_\text{xy}^{t})$
& $10.0$ \\
Z progress
& $c_\text{prog}\sigma_c(d_z^{t-1} - d_z^{t})$
& $10.0$ \\
Landing bonus
& \makecell[l]{$\mathbbm{1}[\text{landed}](r_\text{land} - c_\text{lv}\lVert \mathbf{v}_\text{rel}\rVert$\\
$+\,c_\text{prec}\max(1 - d_\text{xy}/r_\text{lz},0))$}
& $100.0,\;2.0,\;10.0$ \\

\addlinespace
\multicolumn{3}{@{}l}{\textit{Drone-only rewards and penalties}} \\
Crash
& $r_\text{crash}\mathbbm{1}[\text{crash}]$
& $-50.0$ \\
Boundary violation
& $r_\text{bnd}\mathbbm{1}[\text{OOB}]$
& $-50.0$ \\
Descent speed
& $-c_\text{desc}\sigma_c\max(v_\text{desc} - v_\text{max},0)^2$
& $2.0$ \\
Altitude hold
& $-c_\text{alt}(1-\sigma_c)(z_\text{cruise} - z_d)^2$
& $0.3$ \\
Velocity
& $-c_v\max(\lVert \mathbf{v}_d\rVert - v_\text{max}^d,0)^2$
& $2.0$ \\
XY corridor velocity
& $-c_\text{xy}\sigma_c\lVert \mathbf{v}_{d,xy}\rVert^2$
& $5.0$ \\
Attitude
& $-c_\Phi(\phi^2 + \theta^2)$
& $0.01$ \\
Thrust smoothness
& $-c_{\Delta T}(\Delta u_d^T)^2$
& $3.0$ \\
RPY smoothness
& $-c_{\Delta \text{rpy}}\lVert \Delta \mathbf{u}_d^\text{rpy}\rVert^2$
& $5.0$ \\

\addlinespace
\multicolumn{3}{@{}l}{\textit{Rover-only penalties}} \\
Stillness
& $-c_\text{still}\sigma_c v_r^2$
& $0.5$ \\
Yaw rate
& $-c_{\omega}\sigma_c\omega_r^2$
& $0.5$ \\
Lateral/backward
& $-c_\text{lat}(v_{y,r}^2 + \max(-v_{x,r},0)^2)$
& $0.2$ \\
$v_x$ smoothness
& $-c_{\Delta v_x}(\Delta u_r^{v_x})^2$
& $0.03$ \\
$v_y$ smoothness
& $-c_{\Delta v_y}(\Delta u_r^{v_y})^2$
& $0.03$ \\
$\omega_z$ smoothness
& $-c_{\Delta \omega}(\Delta u_r^{\omega_z})^2$
& $0.001$ \\
Rover boundary
& $-c_\text{rb}\mathbbm{1}[\text{rover at edge}]$
& $1.0$ \\
\bottomrule
\end{tabular}
\vspace{2pt}
\begin{flushleft}
\scriptsize
Here $d_\text{xy}=\|[x_d-x_r,\;y_d-y_r]^\top\|_2$ is the horizontal drone-rover distance and
$d_z=|z_d-h_r|$ is the vertical distance to the rover landing-pad height $h_r$.
The corridor gate is
$\sigma_c=\sigma((r_\text{corr}-d_\text{xy})/w)$, with
$r_\text{corr}=0.3\,\text{m}$ and $w=0.1\,\text{m}$.
The relative touchdown velocity is
$\mathbf{v}_\text{rel}=[\dot{x}_d-v_{r,x}^{w},\dot{y}_d-v_{r,y}^{w},\dot{z}_d]^\top$,
$v_\text{desc}=\max(-\dot{z}_d,0)$, and
$\mathbf{v}_{d,xy}=[\dot{x}_d,\dot{y}_d]^\top$.
The rover speed is $v_r=\sqrt{v_{x,r}^2+v_{y,r}^2}$, where
$v_{x,r}$ and $v_{y,r}$ are body-frame velocities, and $\omega_r$ is the rover yaw rate.
The symbols $\phi$ and $\theta$ denote drone roll and pitch.
The terms $\Delta u_d^T$, $\Delta \mathbf{u}_d^\text{rpy}$, and
$\Delta u_r^{(\cdot)}$ denote changes in thrust, drone attitude commands, and rover velocity commands between consecutive time steps.
The constants are $r_\text{lz}=0.07\,\text{m}$, $v_\text{max}=0.3\,\text{m/s}$,
$v_\text{max}^d=0.6\,\text{m/s}$, and $z_\text{cruise}=1.0\,\text{m}$.
The indicator functions denote successful landing, crash/contact failure, drone out-of-bounds (OOB), and rover contact with the arena edge.
The drone receives $r_\text{team}+r_\text{drone}$ and the rover receives $r_\text{team}+r_\text{rover}$.
\end{flushleft}
\end{table}

\begin{table}[t]
\small\sf
\centering
\caption{Curriculum levels and their configurations.}
\label{tab: curriculum landing}
\centering
\setlength{\tabcolsep}{2.5pt}
\scriptsize
\begin{tabular}{@{}c cc cc cc@{}}
\toprule
& \multicolumn{2}{c}{Drone Spawn} & \multicolumn{2}{c}{Rover Spawn} & \multicolumn{2}{c}{Domain Rand.} \\
\cmidrule(lr){2-3} \cmidrule(lr){4-5} \cmidrule(lr){6-7}
Level & $\Delta_{xy}$ & $\Delta_z$ & $v_\mathrm{max}$ & Stationary & $\sigma_m$ & $\sigma_J$ \\
& {[m]} & {[m]} & {[m/s]} & & {[g]} & {[$\!\times\!10^{-6}$]} \\
\midrule
1 & 0.50 & 0.5--1.0 & --- & \checkmark & --- & --- \\
2 & 1.00 & 0.5--1.5 & 0.3 & & --- & --- \\
3 & 1.50 & 0.5--1.5 & 0.6 & & --- & --- \\
4 & 2.50 & 0.5--1.5 & 0.9 & & 2.0 & 1.0 \\
5 & 3.50 & 0.5--1.5 & 0.9 & & 2.0 & 1.0 \\
6 & 4.50 & 0.5--2.0 & 1.0 & & 2.0 & 1.0 \\
\bottomrule
\end{tabular}
\end{table}

\subsubsection{Problem Definition}
To further evaluate MA-AC-MPC, we construct a heterogeneous multi-agent environment where a drone is tasked with landing on a moving ground robot. Each agent receives an observation vector tailored to its role in the cooperative landing task. We denote the drone position and velocity by
\begin{equation}
\mathbf{p}_d = [x_d,y_d,z_d]^\top, \qquad
\mathbf{v}_d = [\dot{x}_d,\dot{y}_d,\dot{z}_d]^\top .
\end{equation}

The relative drone-rover position is
\begin{equation}
\mathbf{p}_{dr} = \mathbf{p}_d - [x_r,y_r,0]^\top,
\end{equation}

with horizontal and vertical distances
\begin{equation}
d_\text{xy} = \|[x_d-x_r,\;y_d-y_r]^\top\|_2, \qquad
d_z = |z_d - h_r|,   
\end{equation}
where $h_r$ is the rover landing-pad height.

For the rover, the observation contains the rover state, relative drone position, and drone velocity:
\begin{equation}
\mathbf{o}_r =
\begin{bmatrix}
\mathbf{x}_r^\top &
\mathbf{p}_{dr}^\top &
\mathbf{v}_d^\top &
\|\mathbf{v}_d\|_2 &
\|\mathbf{p}_{dr}\|_2
\end{bmatrix}^{\top}
\in \mathbb{R}^{15},
\end{equation}
where the rover state is
\begin{equation}
\mathbf{x}_r =
\begin{bmatrix}
x_r & y_r & \cos\theta_r & \sin\theta_r &
v_{x,r} & v_{y,r} & \omega_r
\end{bmatrix}^{\top}
\in \mathbb{R}^{7}.
\end{equation}
Here, $v_{x,r}$ and $v_{y,r}$ are rover body-frame velocities, and $\omega_r$ is the rover yaw rate.
The drone observation contains the drone state, rover motion, and relative position:
\begin{equation}
\mathbf{o}_d =
\begin{bmatrix}
\mathbf{p}_d^\top &
\mathbf{v}_d^\top &
\operatorname{vec}(\mathbf{R}_d)^\top &
\boldsymbol{\omega}_d^\top &
\mathbf{z}_r^\top &
\mathbf{p}_{dr}^\top
\end{bmatrix}^{\top}
\in \mathbb{R}^{29},
\end{equation}
where $\mathbf{R}_d \in SO(3)$ is the drone rotation matrix, $\boldsymbol{\omega}_d$ is the drone angular velocity, and
\begin{equation}
\mathbf{z}_r =
\begin{bmatrix}
x_r & y_r &
(\mathbf{v}_r^{w})^\top &
\sin\theta_r & \cos\theta_r &
v_r &
v_{y,r}
\end{bmatrix}^{\top}
\in \mathbb{R}^{8}.
\end{equation}
The rover speed used in the reward table is
\begin{equation}
v_r = \sqrt{v_{x,r}^2 + v_{y,r}^2},
\end{equation}
and its world-frame velocity is
\begin{equation}
\mathbf{v}_r^{w} =
\begin{bmatrix}
v_{x,r}\cos\theta_r - v_{y,r}\sin\theta_r \\
v_{x,r}\sin\theta_r + v_{y,r}\cos\theta_r
\end{bmatrix}.
\end{equation}
Finally, the relative velocity used in the landing bonus is
\begin{equation}
\mathbf{v}_\text{rel}
=
\begin{bmatrix}
\dot{x}_d - v_{r,x}^{w} &
\dot{y}_d - v_{r,y}^{w} &
\dot{z}_d
\end{bmatrix}^{\top},
\end{equation}

and $\mathbf{v}_{d,xy}=[\dot{x}_d,\dot{y}_d]^\top$. The two agents must cooperate to maximize the cumulative team and individual rewards summarized in \Cref{tab:reward_landing}. 

\begin{figure}[t]
    \centering
    \includegraphics[width=\linewidth]{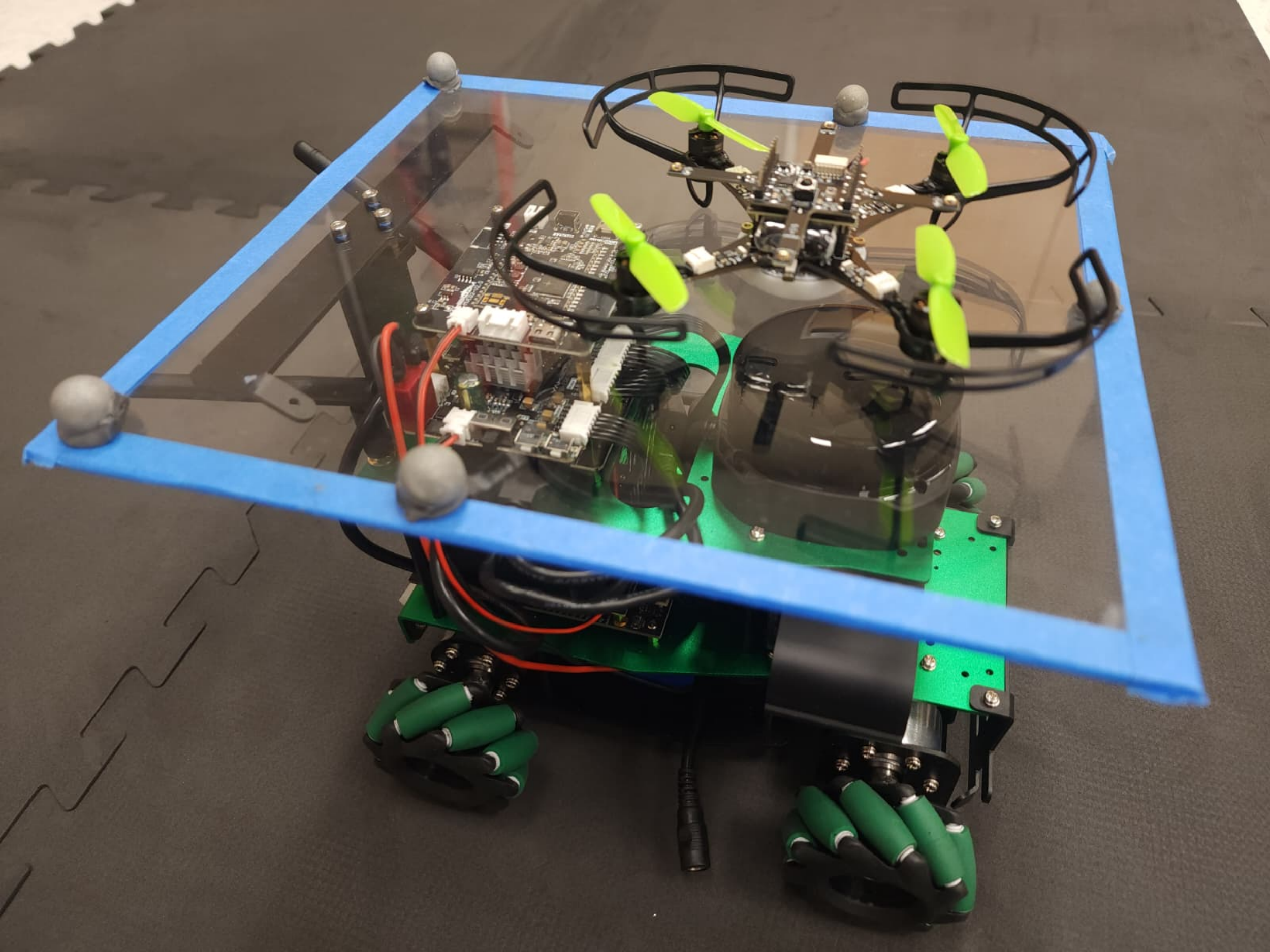}
    \caption{Close-up view of the Crazyflie brushless on top of the ROSMASTER X3 rover.}
    \label{fig:drone_rover_closeup}
\end{figure}

The drone model uses the \verb|first_principles| model for the physics of the environment and \verb|so_rpy| for the MPC model. For the ground robot, we select the Yahboom ROSMASTER X3 shown in \Cref{fig:drone_rover_closeup}. The X3 is a mecanum-wheeled omnidirectional platform that allows independent control of forward, lateral, and rotational velocities. The ground robot uses the following dynamics
\begin{equation}
\begin{aligned}
\dot{x} &= v_{x} \cos\theta - v_{y} \sin\theta \\
\dot{y} &= v_{x} \sin\theta + v_{y} \cos\theta \\
\dot{\theta} &= \omega_{z} \\
\dot{\omega}_{i} &= \frac{\bar{\omega}_{i} - \omega_{i}}{\tau}, \quad i \in \{\mathrm{FL}, \mathrm{FR}, \mathrm{BL}, \mathrm{BR}\}
\end{aligned}
\end{equation}
where $x, y$ are the world-frame position, $\theta$ is the heading, $v_{x}$ and $v_{y}$ are body-frame forward and lateral velocities, $\omega_{z}$ is the yaw rate, $\omega_{i}$ are the individual wheel angular velocities, and $\tau$ is the first-order motor time constant modeling the PID loop response. The body velocities are related to the wheel speeds through the forward kinematics. The target wheel speeds $\bar{\omega}_{i}$ are obtained via the mecanum inverse kinematics
\begin{equation}
\begin{aligned}
\bar{\omega}_{\mathrm{FL}} &= \tfrac{1}{r}\left(v_{x,\mathrm{cmd}} - v_{y,\mathrm{cmd}} - K\omega_{z,\mathrm{cmd}}\right) \\
\bar{\omega}_{\mathrm{FR}} &= \tfrac{1}{r}\left(v_{x,\mathrm{cmd}} + v_{y,\mathrm{cmd}} + K\omega_{z,\mathrm{cmd}}\right) \\
\bar{\omega}_{\mathrm{BL}} &= \tfrac{1}{r}\left(v_{x,\mathrm{cmd}} + v_{y,\mathrm{cmd}} - K\omega_{z,\mathrm{cmd}}\right) \\
\bar{\omega}_{\mathrm{BR}} &= \tfrac{1}{r}\left(v_{x,\mathrm{cmd}} - v_{y,\mathrm{cmd}} + K\omega_{z,\mathrm{cmd}}\right)
\end{aligned}
\end{equation}
with each wheel clipped to $[-\omega_{\max}, \omega_{\max}]$, where $K = l + d$ is the combined kinematic parameter from the standard 45° roller geometry. 
The MPC uses smooth body-level first-order velocity dynamics with state
$\mathbf{x}_r=[x_r,y_r,\theta_r,v_{x,r},v_{y,r},\omega_r]^\top$ and control
$\mathbf{u}_r=[v_{x,\mathrm{cmd}},v_{y,\mathrm{cmd}},\omega_{\mathrm{cmd}}]^\top$ is defined as
\begin{equation}
\begin{aligned}
\dot{x}_r &= v_{x,r}\cos\theta_r - v_{y,r}\sin\theta_r, \\
\dot{y}_r &= v_{x,r}\sin\theta_r + v_{y,r}\cos\theta_r, \\
\dot{\theta}_r &= \omega_r, \\
\dot{v}_{x,r} &= \frac{v_{x,\mathrm{cmd}} - v_{x,r}}{\tau}, \\
\dot{v}_{y,r} &= \frac{v_{y,\mathrm{cmd}} - v_{y,r}}{\tau}, \\
\dot{\omega}_r &= \frac{\omega_{\mathrm{cmd}} - \omega_r}{\tau}.
\end{aligned}
\label{eq:x3_mpc_dynamics}
\end{equation}
This body-velocity input model matches the X3 command interface, which accepts linear and angular velocity commands rather than direct wheel-speed commands. The model is discretized with RK4 inside the MPC. The wheel limits are enforced as linear inequality constraints on the commanded body velocity:
\begin{equation}
-\omega_{\max}\mathbf{1}
\leq
D_{\mathrm{wheel}}\mathbf{u}_r
\leq
\omega_{\max}\mathbf{1},
\end{equation}
where
\begin{equation}
D_{\mathrm{wheel}}
=
\frac{1}{r_w}
\begin{bmatrix}
1 & -1 & -K \\
1 &  1 &  K \\
1 &  1 & -K \\
1 & -1 &  K
\end{bmatrix},
\qquad
K=l+d.
\end{equation}
Parameters are taken from the X3 URDF and ROS2 controller configuration: wheel radius $r_w = 0.0325\,\mathrm{m}$, half-wheelbase $l = 0.08\,\mathrm{m}$, half-track width $d = 0.0845\,\mathrm{m}$, $K = 0.1645\,\mathrm{m}$, motor time constant $\tau = 0.1\,\mathrm{s}$, and maximum wheel speed $\omega_{\max} = 34.9\,\mathrm{rad/s}$. 
Curriculum learning is also used in this multi-agent environment to stabilize learning through a sequence of increasingly difficult parameter configurations shown in \Cref{tab: curriculum landing}.

\begin{figure}[t]
    \centering
    \includegraphics[width=\linewidth]{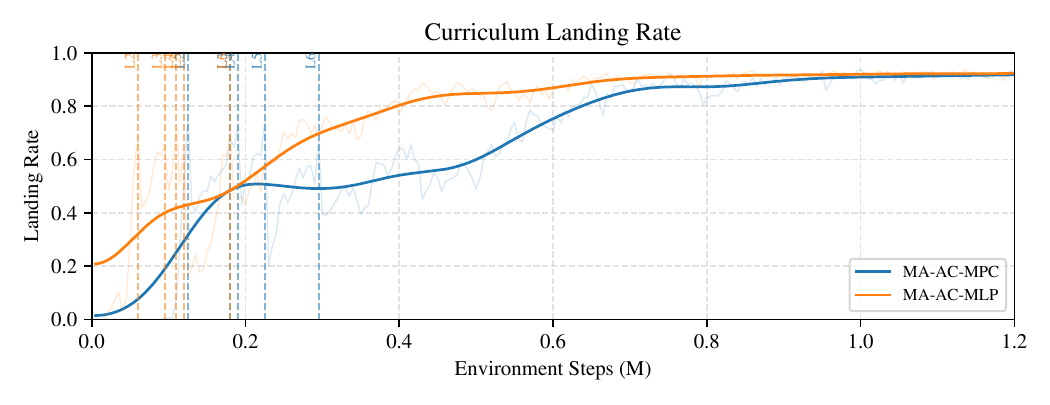}
    \caption{
        Landing rate for MA-AC-MPC and MA-AC-MLP during training.
        Solid curves show the smoothed landing rate, faint curves show the raw logged values,
        and dashed vertical lines indicate curriculum level advancements.
    }
    \label{fig:acmpc_vs_mlp_landing_rate}
\end{figure}

\subsubsection{Simulation and Hardware}
We train both the MA-AC-MPC and MA-AC-MLP with critic and actor network size of $[256,256]$ using ReLU activations. Both methods are trained with MAPPO for $1.2\times10^6$ environment steps using $128$ parallel environments. The PPO rollout length is $256$, with $4$ learning epochs and $4$ mini-batches per update. We use a learning rate of $3\times10^{-4}$, discount factor $\gamma=0.99$, GAE parameter $\lambda=0.95$, clipping ratio $\epsilon=0.2$, entropy coefficient $0.01$, value loss coefficient $1.0$, and gradient norm clipping at $0.5$. The MPC horizon for both drone and rover is $N=2$. We provide plots for the landing rate during training in \Cref{fig:acmpc_vs_mlp_landing_rate}. See Extension 3 for a simulation video.

\begin{figure}[h!]
    \centering
    \includegraphics[width=\columnwidth]{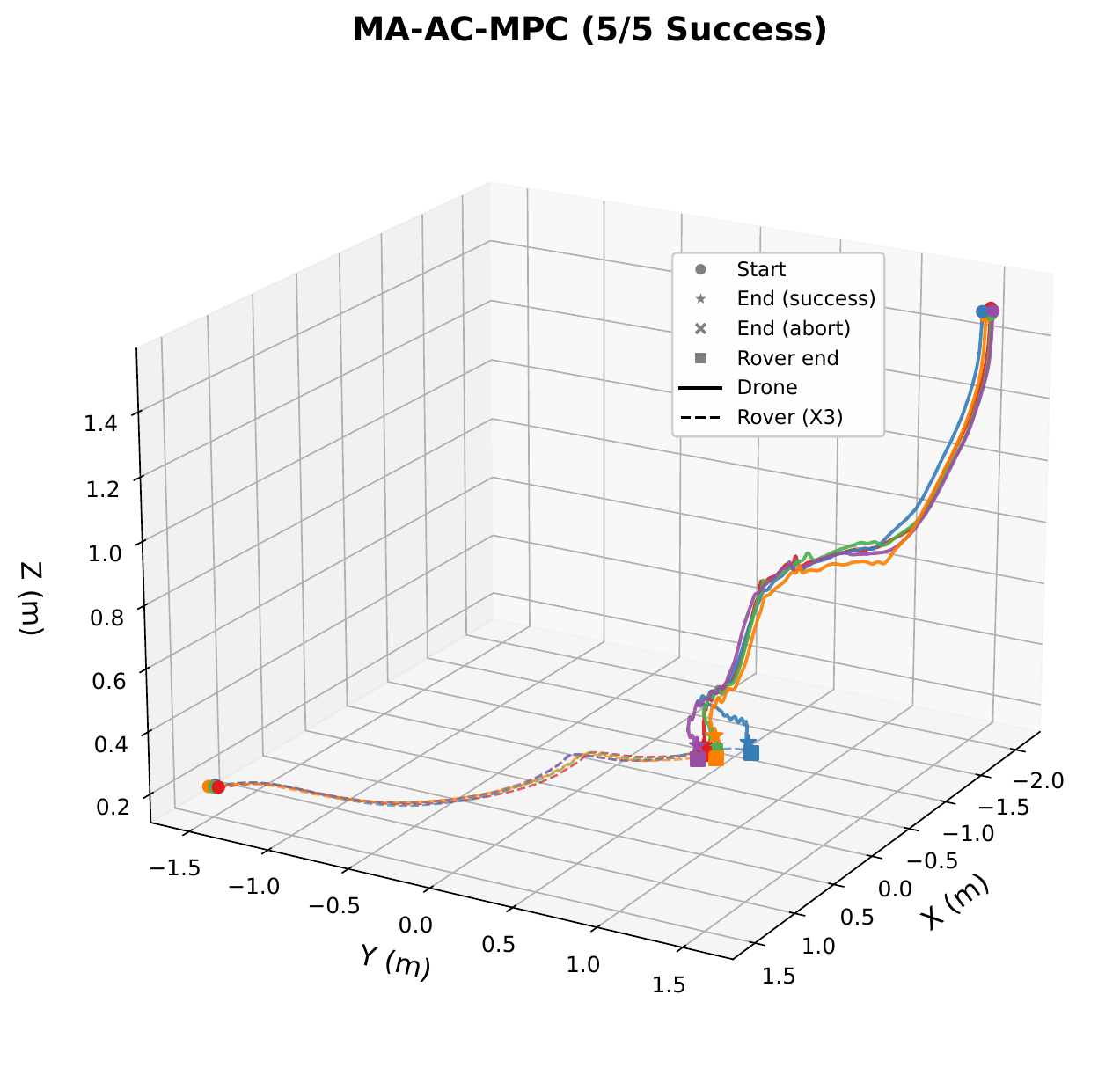}
    
    \vspace{0.5em}
    
    \includegraphics[width=\columnwidth]{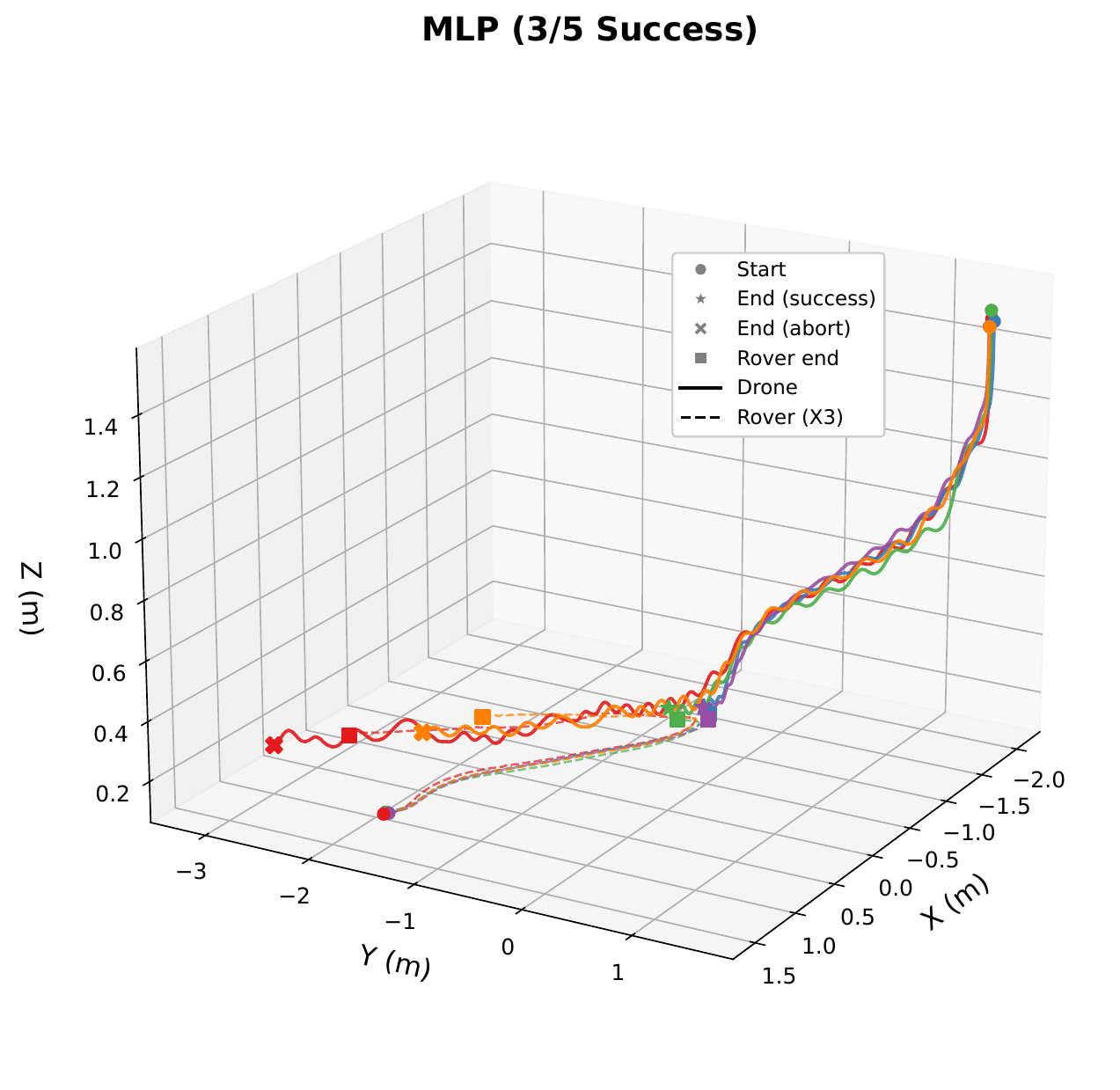}
    \caption{Three-dimensional landing trajectories for the two policies across five hardware trials, with MA-AC-MPC shown on top and MA-AC-MLP on the bottom. In each plot, the drone trajectory is drawn with solid lines and the rover trajectory with dashed lines, while endpoint markers indicate landing outcome. MA-AC-MPC exhibits consistently tight convergence to the rover landing pad across all trials, whereas MA-AC-MLP shows larger variability and two clear failure cases with substantial terminal error. Other runs not shown showed similar behavior for both policies.}
    \label{fig:trajectory_comparison}
\end{figure}

In this particular environment, we noticed that MA-AC-MLP outperforms MA-AC-MPC in sample efficiency during training. However, we tested two trained policies for both the Crazyflie thrust upgrade model \verb|cf2x_T350| and the brushless Crazyflie model in hardware experiments and noticed that MA-AC-MPC consistently landed on the X3 rover every single time with repeatable trajectories while MA-AC-MLP overshot the landing pad and only landed approximately $60\%$ of the time compared to the $100\%$ success rate of MA-AC-MPC. We tabulate results for various hardware trials for both models in \Cref{tab:landing_results}. See Extension 4 and 5 for hardware videos of these trials. We also tested other trials and found consistent results with MA-AC-MPC landing every single time. We also show the trajectories of each trial for both MA-AC-MPC and MA-AC-MLP in \Cref{fig:trajectory_comparison}. This demonstrates that for this environment MA-AC-MPC is more robust and successful at transferring trained policies to hardware than MA-AC-MLP.

\begin{table}[t]
\small\sf\centering
\caption{Landing performance comparison between MA-AC-MPC and MA-AC-MLP.}
\label{tab:landing_results}
\scriptsize
\begin{tabular}{p{0.11\columnwidth} p{0.18\columnwidth} p{0.17\columnwidth} p{0.17\columnwidth} p{0.12\columnwidth}}
\hline
\multicolumn{5}{c}{\textbf{MA-AC-MPC} (Success Rate: 5/5, 100\%, Mean Error: 0.055 m)} \\
\hline
Trial & Result & $\Delta x$ [m] & $\Delta y$ [m] & $d$ [m] \\
\hline
T1 & Success & 0.054 & 0.025 & 0.060 \\
T2 & Success & 0.073 & 0.015 & 0.075 \\
T3 & Success & 0.047 & 0.011 & 0.049 \\
T4 & Success & 0.048 & 0.023 & 0.053 \\
T5 & Success & 0.038 & 0.012 & 0.040 \\
\hline
\multicolumn{5}{c}{\textbf{MA-AC-MLP} (Success Rate: 3/5, 60\%, Mean Error: 0.240 m)} \\
\hline
Trial & Result & $\Delta x$ [m] & $\Delta y$ [m] & $d$ [m] \\
\hline
T1 & Failure & 0.284 & -0.507 & 0.581 \\
T2 & Success & 0.021 & -0.032 & 0.038 \\
T3 & Success & 0.065 & -0.036 & 0.074 \\
T4 & Success & 0.038 & -0.025 & 0.046 \\
T5 & Failure & 0.280 & -0.366 & 0.460 \\
\hline
\end{tabular}
\end{table}


\section{Conclusions}
We propose a framework for combining MARL with model predictive control by extending an existing AC-MPC framework for MARL. Specifically, the MA-AC-MPC framework contains a shared parameter cost network used for the model predictive controller on each agent. In our results, we deploy the framework on a multi-agent pursuit-evasion problem and a drone landing on a rover scenario. We successfully demonstrate higher evader win rates using MA-AC-MPC compared to MA-AC-MLP by sweeping the mass at execution while only being trained on a smaller mass distribution. We also demonstrate improved sample efficiency in training and higher evader win rates during training for MA-AC-MPC while even compared to a larger actor network and critic for MA-AC-MLP. The drone landing on a rover scenario consistently demonstrated a success rate of $100\%$ for MA-AC-MPC in hardware compared to $60\%$ for MA-AC-MLP.


\begin{ethical}
There are no human participants in this article and informed
consent is not required.
\end{ethical}
\begin{contrib}
Christian Llanes: Conceptualization, investigation, methodology, software, writing - original draft. Spencer W. Jensen: Funding acquisition, Supervision. Samuel Coogan: Funding acquisition, project administration, writing – review and editing
\end{contrib}
\begin{acks}
This work received contributions from AI assisted tools for the development of code for simulation and hardware testing of both multi-agent environments using Claude Opus 4.5-4.6. Additionally, Claude Opus was used for generating \Cref{fig: architecture diagram}, code for visualization of data in \Cref{fig:mass sweep mape}, and assisting in developing \Cref{tab: reward mape},\ref{tab: curriculum mape},\ref{tab:reward_landing}, \ref{tab: curriculum landing}, and \ref{tab:landing_results} based on the code and tests from the simulation and hardware. Claude Opus was also used for generating equations 11-25 and sentences to describe the rover dynamics from the environment code. Gemini Pro and Flash 3.1 was used for assisting in grammar checks and improvement in sentence wording. Gemini Pro was also minimally used in pointing to literature articles which were verified by the corresponding author to be relevant to this article.
\end{acks}
\begin{funding}
This work is supported in part by the NASA University Leadership Initiative (ULI) under grant number 80NSSC20M0161. 

This paper describes objective technical results and analysis. Any subjective views or opinions that might be expressed in the paper do not necessarily represent the views of the U.S. Department of Energy or the United States Government. Sandia National Laboratories is a multimission laboratory managed and operated by National Technology \& Engineering Solutions of Sandia, LLC, a wholly owned subsidiary of Honeywell International Inc., for the U.S. Department of Energy’s National Nuclear Security Administration under contract DE-NA0003525. SAND\#0000-XXXXX
\end{funding}
\begin{dci}
The authors declared no potential conflicts of interest with respect to the research, authorship, and/or publication of this article.
\end{dci}

\begin{sm}
The code used to run the simulation and hardware experiments for both multi-agent environments can be found publically available in the links below: \\ 
\href{https://github.com/llanesc/crazyflie-mape-crazyflow}{https://github.com/llanesc/crazyflie-mape-crazyflow} \\ 
\href{https://github.com/llanesc/crazyflie-rover-landing}{https://github.com/llanesc/crazyflie-rover-landing}.
\end{sm}

\bibliographystyle{SageH}
\bibliography{Bibliography.bib} 

\section*{Appendix}
\subsection*{A: Index to multimedia extensions}

\vspace{1ex} 

\noindent
\begin{minipage}{\linewidth}
\small\sf\centering
\captionof{table}{Index of multimedia extensions.}
\label{tab:multimedia_index}
\small
\begin{tabular}{ll>{\raggedright\arraybackslash}p{4.2cm}}
\toprule
\textbf{Extension} & \textbf{\shortstack[l]{Media\\type}} & \textbf{Description} \\ 
\midrule
1 & Video & Simulation of multi-agent pursuit-evasion environment using MA-AC-MPC for evaders. \\ \addlinespace
2 & Video & Hardware demonstratin of multi-agent pursuit-evasion environment using MA-AC-MPC for evaders. \\ \addlinespace
3 & Video & Drone and rover multi-agent environment simulation where both agents use MA-AC-MPC. \\ \addlinespace
4 & Video & Hardware demonstration tests 1 to 6 for drone landing on rover environment using MA-AC-MLP. \\ \addlinespace
5 & Video & Hardware demonstration tests 1 to 5 for drone landing on rover environment using MA-AC-MPC. \\ 
\bottomrule
\end{tabular}
\end{minipage}

\end{document}